\documentclass[12pt]{iopart}

\usepackage{graphicx}
\usepackage{hyperref}
\usepackage{amssymb}
\usepackage{cite}
\usepackage{mathrsfs}
\usepackage{multirow}

\usepackage{color}
\definecolor{greencolor}{rgb}{0,0.5,0.2}
\definecolor{redcolor}{rgb}{0,0.,0.}
\definecolor{bluecolor}{rgb}{0,0.,1.}
\definecolor{greycolor}{rgb}{.5,.5,.5}
\def\Red#1{{\color{redcolor} #1}}

\begin{document}

\title{Probing the topological properties of complex networks modeling short written texts}

\author{Diego R. Amancio}

\address{Department of Computer Sciences \\
Institute of Mathematical and Computer Sciences\\
University of S\~{a}o Paulo, S\~{a}o Carlos, S\~ao Paulo, Brazil}
\ead{diego@icmc.usp.br, diego.raphael@gmail.com}
\vspace{10pt}

\begin{abstract}
In recent years, \Red{graph theory} has been widely employed to probe several language properties. More specifically, the so-called word adjacency model has been proven useful for tackling several practical problems, especially those relying on textual stylistic analysis. The most common approach to treat texts as networks has simply considered either large pieces of texts or entire books. This approach has certainly worked well -- many informative discoveries have been made this way -- but it raises an uncomfortable question: could there be important topological patterns in small pieces of texts? To address this problem, the topological properties of subtexts sampled from entire books was probed. Statistical analyzes performed on a dataset comprising 50 novels revealed that most of the traditional topological measurements are stable for \Red{short subtexts}. 
\Red{
When the performance of the authorship recognition task was analyzed, it was found that a proper sampling yields a discriminability similar to the one found with full texts.
}
Surprisingly, the support vector machine classification based on the characterization of short texts outperformed the one performed with entire books. These findings suggest that a local topological analysis of large documents might improve its global characterization. Most importantly, it was verified, as a proof of principle, that short texts can be analyzed with the methods and concepts of complex networks. As a consequence, the techniques described here can be extended in a straightforward fashion to analyze texts as time-varying complex networks.
\end{abstract}

%
%
%
%
%

\section{Introduction}

\Red{Graph theory} has been employed to probe the statistical properties of many real systems~\cite{app}. Most of the real networks share the small-world~\cite{sw} and scale-free~\cite{sf} properties. The last fifteen years have witnessed the increase of networked models in interdisciplinary applications, including implementations in Physics~\cite{502,509}, Social Sciences~\cite{social1,social2,social3}, Biology~\cite{bio,bio2}, Neuroscience~\cite{olaf2,medidasolaf}, Cognitive Sciences~\cite{cognitive1,cognitive2}, Music~\cite{musique} and Computer Sciences~\cite{mirvezesmardi}. In the latter, graph-based techniques have been applied to the analysis and construction of software architecture~\cite{arch}, supervised classifiers~\cite{highlevel}, spam filters~\cite{spam} and natural language processing (NLP) systems~\cite{congliu}. \Red{For the purpose of textual analysis, networks have proven relevant not only to improve the performance of NLP tasks}~\cite{liu1.0,liu2.0,liu3.0}, but also to better understand the emergent patterns and mechanisms behind the origins of the language~\cite{congliu}.

\Red{Due to its interdisciplinary nature, graph theory can be employed to study the various levels of complexity of the language. In the neuroscience field, the network framework turned out to be a powerful tool for representing the topology of neural systems, where the neocortex is responsible for conveying information~\cite{24}. Among several findings, it has been shown that the clustering and small-world effects facilitate local and global processes, respectively~\cite{cognitive2}. The connection between neuroscience and language/mind processing has been investigated in terms of the topological properties of the connectome~\cite{29}. Interestingly, it has been found that
some linguistic impairments account for fluctuations in the properties of several networks representing brain organization~\cite{francisco,38,28}. At the cognitive level, networks have proven useful to unveil the mechanisms behind information processing~\cite{borges}.
In this context, some studies link certain diseases with specific characteristics of semantic free-association networks~\cite{53}, while other  investigations relate topological network properties with cognitive complexity. For example, the authors in~\cite{68} found that the recognition of a word depends on the average clustering coefficient of the network. Another example is the use of network measurements for quantifying the cognitive complexity of finding the way out of mazes~\cite{labirintoscosta}}.

Network-based models have been applied to study several levels of language organization, which encompasses both the syntactical~\cite{synt1,synt2} and the semantical level~\cite{sem1,sem2,autoCitacaoLiu,lsa1,lsa2,lsa3}. A well-known model is the so-called word adjacency network, which consists in linking adjacent words. Since this model reflects mostly syntactical and stylistic factors~\cite{voynich}, it has been successfully employed for syntactical complexity analysis~\cite{sanda}, detection of literary movements~\cite{literary} and for stylometry~\cite{style,style2}. In most applications, it has been assumed the availability of long texts (or books) to \Red{perform statistical analyses~\cite{highlevel,synt1,voynich,style2,adj,nnw1,nnw2}}. Unfortunately, in many real-world situations, the available of large pieces of texts is uncertain. Whenever only very short pieces of text are available, the conclusions drawn from the analysis could be invalid due to statistical fluctuations present in short written texts. In addition, the unavoidable noise pervading short texts could undermine the performance of NLP tasks. Therefore, it becomes of paramount importance to know beforehand if a given text is long enough for the analysis. In this context, this paper investigates how the selection of short pieces of texts (hereafter referred to as samples) affects the topological analysis of word adjacency networks.

%

\Red{In order to study the fluctuations of networks measurements modeling short texts, books were sampled in adjacent, non-overlapping subtexts. As I shall show,
the sampling of texts yields subtexts with similar topology, as revealed by a systematic analysis of the variability of several measurements across distinct samples.} The influence of the subtext length on the authorship recognition task was also studied. The results revealed that the best performance was achieved when the books were split in shorter subtexts, which confirms that the sampling might favor the classification process as the local topological characterization of books becomes more precise.

\section*{Materials and Methods}

In this section, the word adjacency model is presented. In addition, I swiftly describe the measurements employed for characterizing networks and the methods used for recognizing topological patterns.

\subsection*{Modeling texts as complex networks}

\Red{
The overall purpose of this paper is to study features of the language
that reflect particular choices made by individuals or groups. Such particular choices, referred to as stylistic features, can be employed e.g. to classify genres, dialects and literary works~\cite{congliu}. A traditional stylistic feature used e.g. to identify authorship is the frequency of a particular function word in a text~\cite{surveymetodos}. Here, a network model is used to capture particular connectivity patterns that might be useful to identify authorship, genres, languages etc.}

%
%

\Red{
There are several ways to model texts as complex networks~\cite{graph-based}. The most appropriate modeling depends on the target application. One of the most employed models for grasping stylistic features is the so-called co-occurrence (or adjacency network)~\cite{voynich,adj}}. Besides capturing \Red{syntactical} attributes of the texts, \Red{networked models} have also proven useful to capture language dependent features~\cite{voynich}.
\Red{
Co-occurrence networks have been employed in many applications~\cite{graph-based}. In generic terms, a co-occurrence network can be defined in a manifold way: two words are linked if they co-occur in at least one window. The window can be represented as n-grams, sentences, paragraphs or even entire documents~\cite{wsdsurvey}. Some alternative co-occurrence models link two words only if the co-occurrence frequency exceeds a given threshold. There is also the possibility to improve the model by including weighted links~\cite{ted}.
Statistical analysis revealed that most word co-occurrence networks display both small-world and scale-free behaviors~\cite{swhl}. The average clustering coefficient and the average nearest neighbor degrees were found to follow a power-law distribution, as a consequence of the existence of distinct functional classes of words~\cite{macci}. Particularly, it has been shown adjacency and syntactic networks display similar topological properties, as far as topological attributes are considered~\cite{synt1}. Besides being useful for analyzing styles in texts, co-occurrence graphs serve to model semantical relationships~\cite{hyperlex,agraphmodel}.
}

Prior to the transformation of the text into a network model, some pre-processing steps are usually applied. Firstly, words conveying low semantic content (such as articles and prepositions) are removed. These words, referred to as \emph{stopwords}, are disregarded from the analysis because they simply serve to connect content words. \Red{Even though previous studies used the frequency of stopwords to classify texts according to their styles, I decided not to use them because I am interested in the relationships between words with significant semantic content. This procedure has been applied in many studies (see e.g.~\cite{antiqueira,style,macci,comparing})}.
In the next step, each word is transformed into its canonical form so that conjugated verbs and nouns are respectively mapped to their infinitive and singular forms. As such, distinct forms of the same word are mapped to the same concept. To obtain the canonical form of words, it is necessary to perform a sense disambiguation~\cite{wsdsurvey} at the word level. To assist the disambiguation process, the part-of-speech of each word is inferred from a maximum entropy model~\cite{ratna}.

\begin{figure}[!htbp]
\begin{center}
    \includegraphics[width=0.5\linewidth]{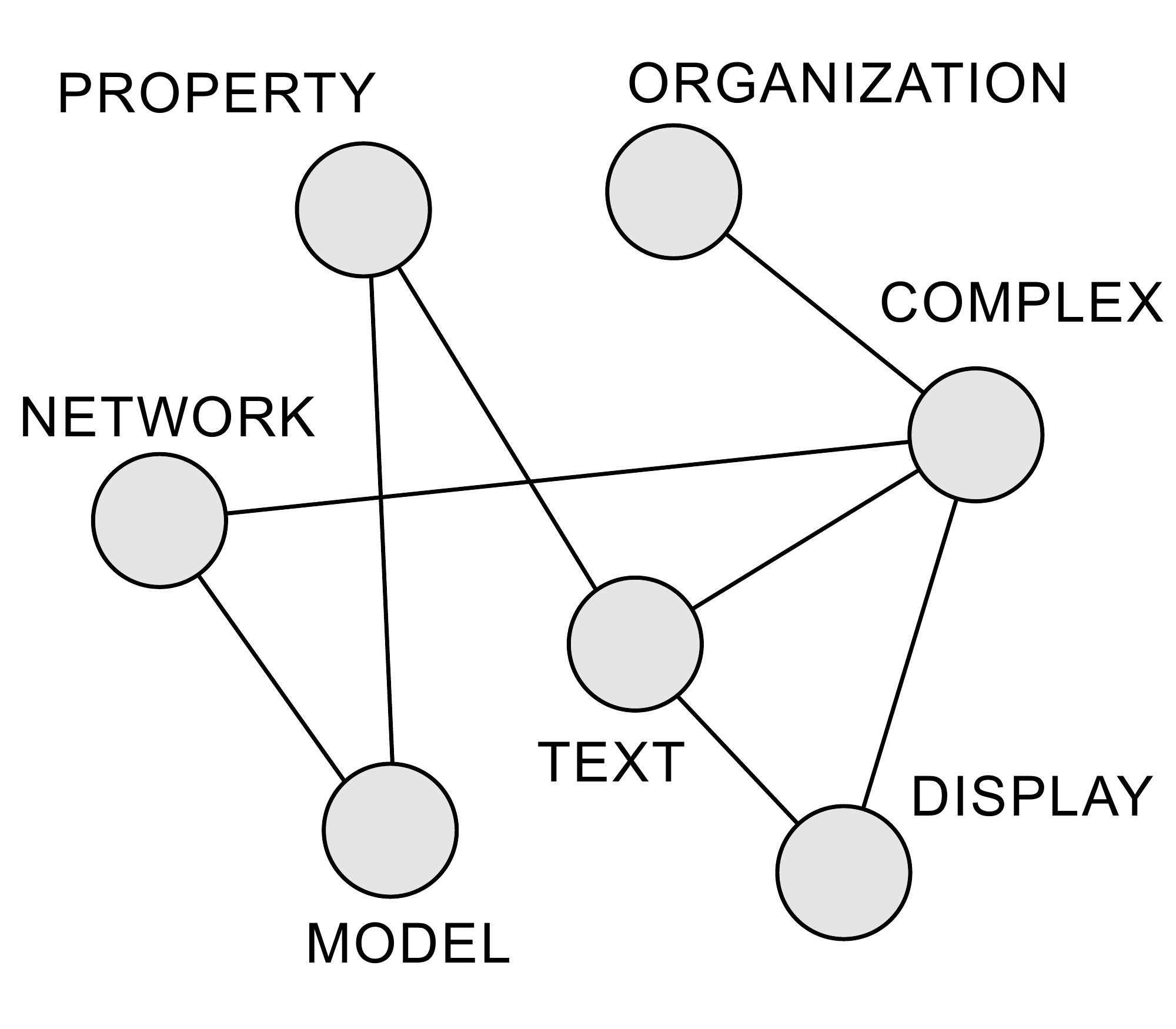}
        \caption{\label{f:redinhaexemplo}Example of adjacency network created from the extract: ``Complex networks model several properties of texts. A complex text displays a complex organization''. After the pre-processing step, words are mapped into nodes, which in turn are connected if the respective words appeared at least once as neighbors.    }
	\end{center}
\end{figure}

After the pre-processing step, each distinct word remaining in the text becomes a node in the network. Therefore, the total number of nodes will be given by the vocabulary size of the pre-processed text. Edges linking two words are created if these words appeared as neighbors in the text at least once. \Red{For example, in the short sentence ``\emph{Complex network measurement}'', the following links are created: \emph{complex} $\rightarrow$ \emph{network} and  \emph{network} $\rightarrow$ \emph{measurement}. In this case, \emph{complex} and \emph{measurement} are not connected to each other because they are separated by one intermediary word. Particularly, this model was chosen here because it has been successfully used in applications where authors' styles represent an important feature for analyzing written texts~\cite{swhl,interplay}.}
Table \ref{tab.pre} and Figure~\ref{f:redinhaexemplo} illustrate the creation of a word adjacency network.

\begin{table*}[h]
\begin{center}
\caption{\label{tab.pre}Pre-processing steps.}
\begin{tabular}{ccc}
\hline
{\bf Orignal Text} & {\bf After pre-processing}  & {\bf Canonical form} \\
\hline
Complex networks model & complex networks model & complex network model \\
several properties of  & properties & property \\
texts. A complex text  & texts complex text & text complex text \\
displays a complex     & displays complex & display complex\\
organization.          & organization & organization \\
\hline
\end{tabular}
\end{center}
\centering
{Example of pre-processing steps performed in a extract for the purpose of creating a adjacency network. The pre-processing step eliminates punctuation marks and words conveying low semantic content. The lemmatization step aims at mapping each word to its canonical form.}
\end{table*}



\subsection*{Topological characterization of textual networks}

The topological analysis of complex networks can be conducted through the use of a myriad of measurements. In the current study, we employed the main measurements that have been used in the analysis of word adjacency networks.
\Red{The quasi-local topology, which considers the connectivity of neighbors, neighbors of neighbors and further hierarchies~\cite{refDoPipinho}, was measured with the clustering coefficient ($C$) and with the average neighbor degree ($k_n$).}
Both measurements have been employed to identify words appearing in generic contexts~\cite{comparing}. In addition to these measurements, the accessibility ($\alpha$) was used to characterize the quasi-local structure of textual networks. To define this measurement, consider that $P_{ij}^{(h)}$ represents the probability of a random walker starting at node $v_i$ to reach node $v_j$ in $h$ steps. Mathematically, the accessibility $\alpha_i^{(h)}$ is defined as the entropy of the quantity $P_{ij}^{(h)}$:
\begin{equation}
    \alpha_i^{(h)} =  \exp \Big{(}-\sum P_{ij}^{(h)}  \ln P_{ij}^{(h) } \Big{)}.
\end{equation}
It has been show that, when the walker performs a self-avoiding random walk, the accessibility is useful to identify the core of textual networks~\cite{key}. Moreover, this measurement has proven useful to generate informative extractive summaries~\cite{synt2}.

In addition to the quasi-local measurements, we also employed some global measurements. The average shortest path length ($l$) was used because this measurement has been useful in several textual applications~\cite{comparing}. Usually, the words taking the lowest values of $l$ are either keywords or words that appear near to the most relevant words in the text.
Shortest paths were also employed in the so called betweenness ($B$)~\cite{btw}. In textual networks, the betweenness grasps the amount of contexts in which a given word may occur~\cite{comparing}. Differently from the clustering coefficient, the betweenness uses the global information of the network to infer the quantity of semantic contexts in which a word occurs.

%
\Red{The spatial distribution of words along the text was studied in terms of the intermittency, a measurement that is is able to capture the irregularity (bursts) of the distribution~\cite{statistical}}. To compute the intermittency, one represents the pre-processed text as a time series. As a consequence, the first word is the first element of the time series, the second word is the second element and so forth. The recurrence time  $t_i$ of a word $w$ is defined as the number of words between two successive occurrences of $w$.
Therefore, a word occurring $N$ times in the text generates the sequence $T = \{t_1, t_2, ..., t_{N - 1} \}$ of recurrence times. In order to consider the time $t_I$ until the first occurrence of the word and the time $t_F$ after the last occurrence, this measurement considers also the time
$t_{N} = t_I + t_F$ in $T$.
If  $\langle t \rangle$ is the mean of the elements in $T$, then the intermittency $I$ of the distribution $T$ is
\begin{equation}
    I = \Bigg{[}{ \frac{ \langle t^2 \rangle}{\langle t \rangle^2} - 1 }\Bigg{]}^{1/2}.
\end{equation}
The intermittency has been employed to identify keywords using features not correlated with word frequencies~\cite{statistical}. Moreover, this measurement has proven useful to classify texts according to the informativeness criterium~\cite{voynich}.

\subsection*{Pattern recognition methods}

In this study, we investigated if the topology of small pieces of texts is able to provide relevant attributes for textual analysis.
\Red{To quantify the effects of sampling large texts in applications based upon the classification of distinct styles, supervised pattern recognition methods were used~\cite{duda}}.
%
%
In a supervised classification problem, we are given a \emph{training set} and a \emph{test set}. The training set $\mathcal{S}_{{tr}} = \{ \beta_{(tr,1)}, \beta_{(tr,2)},\ldots, \beta_{(tr,n)} \}$ is the set of examples that inference algorithms employ to generate classification models. After the creation of the classification model, the test dataset $\mathcal{S}_{{ts}} = \{ \beta_{(ts,1)}, \beta_{(ts,2)},\ldots, \beta_{(ts,m)}  \}$  is employed to
evaluate the classification performance. The result of the classification is the mapping $\mathcal{S}_{{ts}} \mapsto \mathcal{C} = \{c_1,c_2,...\}$. In other words, a conventional classifier assigns a unique class $c_i \in \mathcal{C}$ for each element of the training set.
For each example $\beta$, the value of the attribute $F_i$ taken by $\beta$ is represented as $\beta^{(i)}$.
The performance of the classification was verified with the well known 10-fold cross-validation technique~\cite{duda}.

The supervised classifiers employed in this paper were: nearest neighbors (kNN)~\cite{nearest}, decision trees (C4.5)~\cite{trees}, bayesian decision (Bayes)~\cite{idiot} and support vector machines (SVM)~\cite{suportes}. Below I present a swift description of these methods. Further details can be found in~\cite{duda}.

\begin{description}

  \item {\bf Nearest neighbors}: this technique classifies a new example $\beta \in \mathcal{S}_{{ts}}$ according to a voting process performed on $\mathcal{S}_{{tr}}$. If most of the $\kappa$ nearest neighbors of $\beta$ belongs to the class $c_i \in \mathcal{C}$, then the class $c_i$ is associated to $\beta_{ts}$.

      \item {\bf Bayesian decision}: this method computes the probability $P(c_i|\beta)$ that a given class $c_i \in \mathcal{C}$ is the correct class associated to a given instance $\beta_{(ts)}$. Assuming that the attributes are independent, $P(c_i|\beta)$ can be computed as
          \begin{equation}
            P(c_i|\beta) = \frac{P(c_i)}{P(F_1 = \beta^{(1)},\ldots)} \prod_{k} P(F_k = \beta^{(k)} | c_i ).
          \end{equation}
          Therefore, the correct class $c_\beta$  is
          \begin{equation}
            c_\beta = \arg \max_{c_i \in \mathcal{C} }  P(c_i)   \prod_{k} P(F_k = \beta^{(k)} | c_i ).
          \end{equation}

       \item {\bf Decision trees}: this algorithm is based upon the induction of a tree, a widely employed abstract data type. To construct a tree model, it is necessary to find the most informative attribute, i.e. the attribute that provides the best discriminability of the data. To do so, several measurements have been proposed~\cite{duda}. In this paper, we use the information gain $\Omega$, which is mathematically defined as
           \begin{equation}
            \Omega( \mathcal{S}_{tr}, F_k ) = \mathcal{H}(\mathcal{S}_{tr}) - \mathcal{H}(\mathcal{S}_{tr}|F_k),
           \end{equation}
           where  $\mathcal{H}(\mathcal{S}_{tr})$ is the entropy of the dataset $\mathcal{S}_{tr}$ and $\mathcal{H}(\mathcal{S}_{tr}|F_k)$ is the entropy of the dataset when the value of $F_k$ is specified. $\mathcal{H}(\mathcal{S}_{tr}|F_k)$ can be computed from the training dataset as
           \begin{eqnarray}
                 \mathcal{H}(\mathcal{S}_{tr}|F_k) & =   \sum_{v \in V(F_k)} &  \frac{ | \beta_{(tr)} \in \mathcal{S}_{tr} |  \beta_{(tr)}^{(k)} = v | }{ | \mathcal{S}_{tr} | } \cdot \nonumber \\
                 &  \mathcal{H}( \{  \beta_{(tr)} \in \mathcal{S}_{tr} |  \beta_{(tr)}^{(k)} = v  \},
           \end{eqnarray}
           where $V(F_k)$ is the set of all values taken by the attribute $F_k$ in the training dataset, i.e.
           \begin{equation}
                V(F_k) =  \bigcup_{i=1}^{|\mathcal{S}_{tr}|} \beta_{(tr,i)}^{(k)}.
           \end{equation}

           \item {\bf Support Vector Machines}: this technique
               divides the attribute space using hyperplanes, so that each region is assigned to a single class. The construction of the hyperplanes relies upon the definition of linear or non-linear kernel functions.
               Once the separation is determined, a new example can be classified by evaluating its position on the attribute space. This method has been applied in several real applications due to its robustness with regard to the number of dimensions and other features~\cite{top10}.


\end{description}

\section*{Results}

\subsection*{Variability of measurements}

In this paper, a set of subtexts sampled from a entire book is considered as consistent when the measurements computed for the subtexts display low variability across different subtexts. I take the view that authors tend to keep their styles across distinct portions of the same book.
This assumption is reasonable because it has been shown that the main factors responsible for stylistic variations in texts are the language~\cite{voynich}, the authorship~\cite{comparing}, the complexity~\cite{sanda} and the publication date~\cite{literary}. As such, it is natural to expect low variability across distinct samples since all these factors remain constant in the same book. Hence, I consider that the main factor accounting for the variability of the style across distinct parts of the same book is the sample size.

In order to compute the variability of the measurements across distinct subtexts, the following procedure was adopted. A dataset comprising
 50 novels (see Tables \ref{tab.books.variance.1} and \ref{tab.books.variance.2}) was used. Each book was split in subtexts comprising $W$ tokens. If one considers a book as a sequence of tokens $\mathcal{W} =  \{w_1,w_2,\ldots\}$, the subtext $\mathbb{T}_i$ will contain the sequence $ \{ w_ {\textrm{S}_i} ,w_{\textrm{S}_i+1}, \ldots,w_{\textrm{S}_i+W} \}$, where $\textrm{S}_i = W \cdot i+1$ and $i \in \mathbb{N}$. The variability of a given measurement $X$ across distinct subtexts $\mathbb{T}_i$'s of a given full book will be given by the coefficient of variation
\begin{equation}
    \nu(X) = \Bigg{[}{ \frac{ \langle X^2 \rangle}{\langle X \rangle^2} - 1 }\Bigg{]}^{1/2}.
\end{equation}
The variability of the following measurements were investigated in the current paper
\begin{equation}
 X = \{\langle \alpha^{(h=2)} \rangle, \langle \alpha^{(h=3)} \rangle, \langle k_n \rangle, \langle B \rangle, \langle C \rangle, \nonumber
\end{equation}
\begin{equation}
\langle I \rangle, \Delta I, \gamma(I), \langle l \rangle, \Delta l \textrm{ and } \gamma(l) \}, \nonumber
\end{equation}
where $\langle \ldots \rangle$, $\Delta$ and $\gamma$ represent the mean, the standard deviation and the skewness of the distribution of the measurements in a given subtext.
\Red{
The accessibility was used to measure the prominence of a given word considering its nearest concentric neighborhood ($h=2$ and $h=3$). Higher values of $h$ were not employed because the accessibility computed in higher levels is not informative~\cite{refDoPipinho}. The clustering coefficient and the average nearest neighbors degrees were used to quantify the connectivity between neighbors. The global prominence was measured with the average shortest path lengths and betweenness. Finally, the intermittency was employed to quantify the relevance of the words according to their spatial distribution along the text.
}

\begin{table*}[h]
\begin{center}
\caption{\label{tab.books.variance.1}List of Books (part 1).}
\begin{tabular}{cll}
\hline
{\bf Date} & \multirow{1}{*}{\bf Author}  & \multirow{1}{*}{\bf Book} \\
\hline
1811 & Jane Austen & Sense and Sensibility\\
1815 & Jane Austen & Emma \\
1826 & James F. Cooper & The Last of the Mohicans\\
1841 & Charles Dickens & Barnaby Rudge: A Tale of the Riots of Eighty\\
1842 & Charles Darwin & The Structure and Distribution of Coral Reefs\\
1842 & Charles Dickens & American Notes for General Circulation\\
1844 & Charles Darwin & Geological Observations on the Volcanic Islands\\
1844 & Charles Darwin & Geological Observations on the South America\\
1847 & Charlotte Bronte & Jane Eyre \\
1847 & William M. Thackeray  & Vanity Fair: A Novel without a Hero\\
1847 & Emily Bronte & Wuthering Heights \\
1850 & Charles Dickens & David Copperfield\\
1851 & Herman Melville & Moby-Dick; or, The Whale\\
1854 & Charles Dickens & Hard Times -- For These Times\\
1856 & Gustave Flaubert & Madame Bovary \\
1859 & Charles Dickens & A Tale of Two Cities \\
1859 & Wilkie Collins & Woman in White\\
1861 & Charles Dickens & Great Expectations \\
1868 & Louisa May Alcott & Little Women\\
1869 & Mark Twain & The Innocents Abroad\\
1869 & Leo Tolstoy & War and Peace \\
1872 & Charles Darwin & The Expression of the Emotions in Man and Animals\\
1873 & Thomas Hardy & A Pair of Blue Eyes\\
1874 & Thomas Hardy & Far From the Madding Crowd\\
1876 & Thomas Hardy & The Hand of Ethelberta: A Comedy in Chapters\\
\hline
\end{tabular}
\end{center}
\centering
{List of books employed for the analysis of variability of complex network measurements.}
\end{table*}

\begin{table*}[h]
\begin{center}
\caption{\label{tab.books.variance.2}List of Books (part 2).}
\begin{tabular}{cll}
\hline
{\bf Date} & \multirow{1}{*}{\bf Author}  & \multirow{1}{*}{\bf Book} \\
\hline
1876 & George Eliot & Daniel Deronda \\
1877 & Leo Tolstoy & Anna Karenina \\
1877 & Charles Darwin & The Different Forms of Flowers on Plants of the Same Species\\
1883 & Mark Twain & Life on the Mississippi\\
1884 & Mark Twain & Adventures of Huckleberry Finn\\
1886 & Thomas Hardy & The Mayor of Casterbridge\\
1887 & Arthur Conan Doyle & A Study in Scarlet \\
1895 & Thomas Hardy & Jude the Obscure \\
1897 & Arthur Conan Doyle & Uncle Bernac \\
1900 & Arthur Conan Doyle & War in South Africa\\
1903 & Bram Stoker & The Jewel of Seven Stars \\
1905 & Bram Stoker & The Man \\
1909 & Bram Stoker & The Lady of the Shroud\\
1911 & Bram Stoker & The Lair of the White Worm \\
1912 & Arthur Conan Doyle & The Lost World\\
1914 & Bram Stoker &  Dracula's Guest \\
1914 & Arthur Conan Doyle & The Valley of Fear \\
1915 & P. G. Wodehouse & Something New \\
1915 & Virginia Woolf & The Voyage Out \\
1920 & Edith Wharton & The Age of Innocence \\
1921 & P. G. Wodehouse & The Girl on the Boat \\
1921 & P. G. Wodehouse & Indiscretions of Archie \\
1922 & P. G. Wodehouse & The Adventures of Sally\\
1922 & P. G. Wodehouse & The Clicking of Cuthbert \\
\hline
\end{tabular}
\end{center}
\centering
{List of books employed for the analysis of variability of complex network measurements.}
\end{table*}

\begin{figure*}[!htbp]
\begin{center}
    \includegraphics[width=0.93\linewidth]{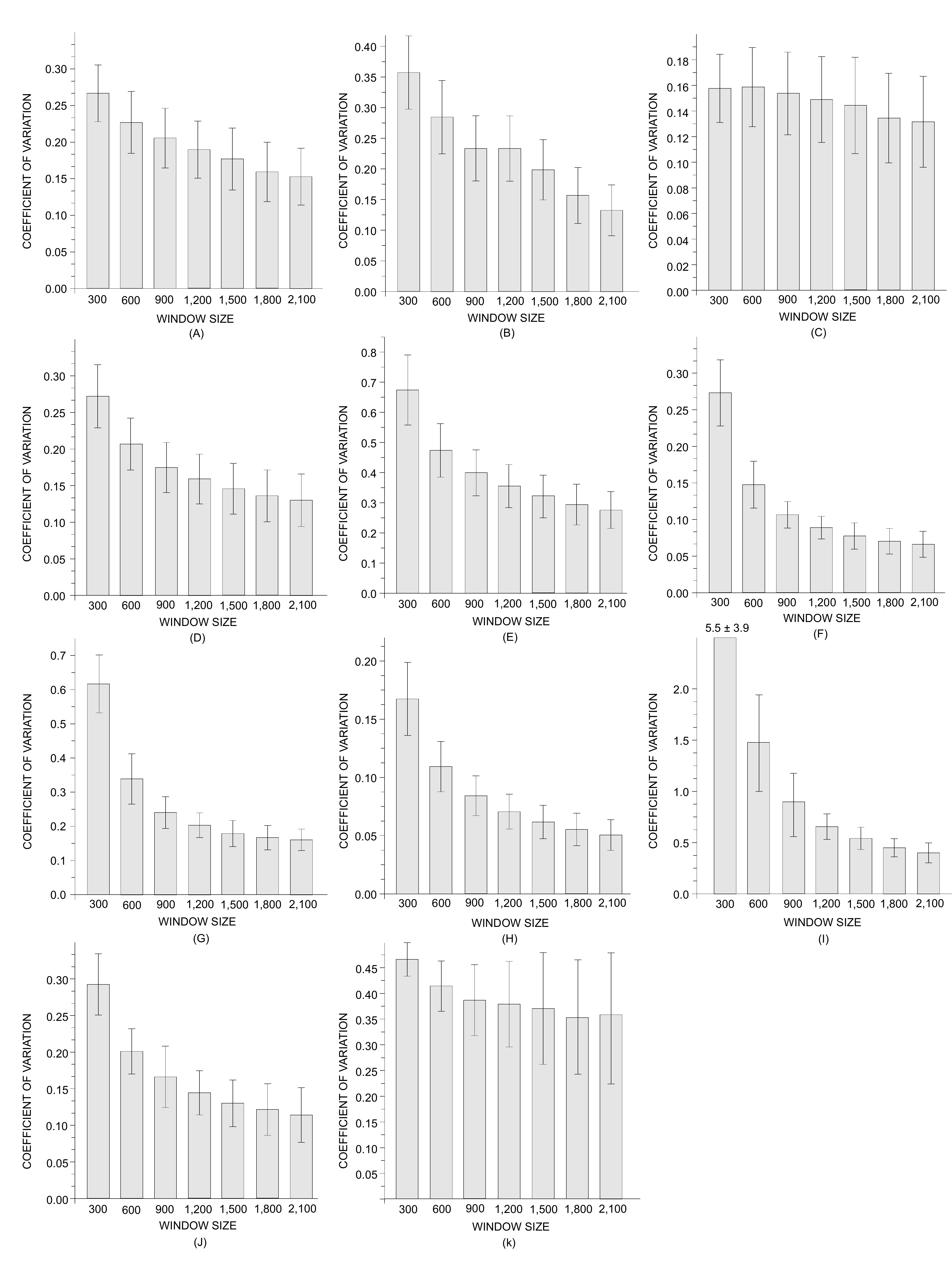}
        \caption{\label{f:vary_num_features} Coefficient of variations for the measurements (a) average acessibility $\langle \alpha^{(h=2)} \rangle$; (b) average acessibility $\langle \alpha^{(h=3)} \rangle$; (c) average neighbors degree $\langle k_n \rangle$; (d) average betweenness $\langle B \rangle$; (e) average clustering coefficient $\langle C \rangle$; (f) average intermittency $\langle I \rangle$; (g) standard deviation of the intermittency $\Delta I$; (h) skewness of the intermittency $\gamma(I)$; (i) average shortest path length $\langle l \rangle$; (j) standard deviation of the shortest path length $\Delta l$; and (k) skewness of the intermittency $\gamma(I).$  }
	\end{center}
\end{figure*}

The variability obtained for each $X$ across distinct subtexts is shown in Figure \ref{f:vary_num_features}. The results confirm the the variability $\nu(X)$ of all measurements studied shows the same behavior, as revealed by a decreasing tendency as $W$ increases. This means that the statistical fluctuations across distinct subtexts decrease as larger portions of subtexts are considered.
The majority of the measurements displayed a variability below $0.35$ for $W \geq 1,500$. The average accessibility displayed a typical coefficient of variation below $0.20$ for  $W \geq 1,500$. The average neighbor degree turned out to be the measurement taking the lowest values of variability.
Even when very small pieces of texts were taken into account (W=300), the typical variability was always below 0,20. The average betweenness also took low values of variability for small subtexts.
However, the lowest values were found for $W \geq 1,500$. The average clustering coefficient was one of the measurements whose variability across subtexts displayed a high dependence upon $W$. More specifically, for small subtexts, $\langle C \rangle$ turned out to be unstable, as revealed by coefficient of variations surpassing $\nu = 0.65$. This result shows that $\langle C \rangle$ should not be employed for the topological analysis of small texts because it is very sensitive to the sampling size. Concerning the intermittency, both $\langle I \rangle$ and $\Delta I$ displayed low values of variability for $W \geq 600$. Conversely, the skewness $\gamma(I)$ displayed high variabilities even for large texts ($W=2,100$). This result might be a consequence of the fact that $\gamma(I)$ reflects the fraction of keywords in a text~\cite{comparing}. Therefore, if the amount of relevant words in each subtext presents a high variability, it is natural to expect that
$\gamma(I)$ will vary accordingly. With regard to the shortest path lengths, both $\langle l \rangle$ and $\Delta l$ displayed low values of fluctuations for $W \geq 1,500$. Similarly to $\gamma(I)$, $\gamma(l)$ presented a high dispersion across distinct subtexts.

All in all, the results displayed in Figure \ref{f:vary_num_features} reveal the most of measurements displays low statistical fluctuations when shorter texts are analyzed.
The only exceptions were the skewness of the average shortest path lengths and the skewness of the intermittency. In both cases, the variability remained high even for subtexts comprising $2,100$ tokens. In any case, in general, it is reasonable to suppose that a proper sampling allows a proper characterization of the \emph{local} topological properties of books. \Red{In order to verify the applicability of sampling books in real stylometric tasks, the next section investigates how the sampling affects the performance of the authorship recognition task~\cite{aa}}.

\subsection*{Authorship recognition via topological analysis of subtexts}

\Red{
Authorship recognition methods are important because they can be applied e.g to solve copyright disagreements~\cite{granted}, to intercept terrorist messages~\cite{abbasi} and to classify literary manuscripts~\cite{ebra}. Automatic authorship recognition techniques became popular after the famous investigation of Mosteller and Wallace on the Federalist Papers~\cite{Mosteller}. After this seminal study, researchers have tried to discover novel features to quantify styles, i.e. the textual properties that unequivocally identify authors. Currently, this line of research is known as stylometry~\cite{aa}. Traditional features employed to discriminate authors include statistical properties of words (average length, frequency and intermittency of specific words, richness of vocabulary)~\cite{surveymetodos} and characters (frequency and correlations)~\cite{granted}. In addition to the lexical features, syntactical (frequency of specific parts-of-speech or chunks) and semantical properties (semantical dependencies) have been employed~\cite{surveymetodos}. Current research have devised new attributes for the development of robust classifiers~\cite{surveymetodos}. In recent years, it has been shown that the topological properties of complex networks has been useful to capture various textual attributes related to authors' styles~\cite{congliu,sanda,literary,style,style2,adj}. For example, a significant discriminability of authors could be found in~\cite{comparing}. Because authors leave stylistic marks on the topological structure of complex networks~\cite{comparing,adj,antiqueira}, in this section the authorship recognition problem is studied by measuring the topological properties of co-occurrence networks. More specifically, the effects of the sampling on the classification are investigated.}


%
\Red{For the classification task, the list of books and authors considered is shown in Table~\ref{tab.books.recog}. Subtext lengths ranging in the interval $500 \leq W \leq 21,400$ were considered. To evaluate the performance of the task in short texts, note that low values of $W$ were used. A comparison of the performance obtained with short and large texts was carried out by analyzing the accuracy rates found for higher values of $W$. In the dataset shown in Table~\ref{tab.books.recog}, the larger window considered ($W=21,400$) corresponds to the case where no sampling was performed. In other words, when $W=21,400$ each book was represented by a single subtext. The values of $W = 7,130$ and $W=5,350$ correspond to the division of full texts in three and four parts, respectively.} Four classifiers were employed to perform the supervised classifications: (i) nearest neighbors, (ii) naive Bayes, (iii) decision trees, and (iv) support vector machines.

\begin{table*}[h]
\begin{center}
\caption{\label{tab.books.recog}List of books employed for the authorship recognitions task.}
\begin{tabular}{cll}
\hline
{\bf Date} & \multirow{1}{*}{\bf Author}  & \multirow{1}{*}{\bf Book} \\
\hline
1892 & Arthur C. Doyle & The Adventures of Sherlock Holmes \\
1907 & Arthur C. Doyle & Through the Magic Door \\
1898 & Arthur C. Doyle & The Tragedy of Korosko \\
1915 & Arthur C. Doyle & The Valley of Fear \\
1902 & Arthur C. Doyle & The War in South Africa\\
\hline
1914 &  Bram Stoker & Dracula's Guest \\
1903 &  Bram Stoker & The Jewel of Seven Stars \\
1909 &  Bram Stoker & The Lady of the Shroud \\
1911 &  Bram Stoker & The Lair of the White Worm \\
1905 &  Bram Stoker & The Man\\
\hline
1842 & Charles Darwin & The Structure and Distribution of Coral Reefs\\
1877 & Charles Darwin & The Different Forms of Flowers\\
1872 & Charles Darwin & The Expression of the Emotions in Man and Animals\\
1844 & Charles Darwin & Geological Observations on the Volcanic Islands\\
1844 & Charles Darwin & Geological Observations on the South America\\
\hline
1914 & Hector H. Munro & Beasts and Super Beasts \\
1911 & Hector H. Munro & The Chronicles of Clovis \\
1919 & Hector H. Munro & Toys of Peace \\
1912 & Hector H. Munro & The Unbearable Bassington\\
1913 & Hector H. Munro & When William Came \\
\hline
\end{tabular}
\end{center}
\centering
{List of books employed for the authorship recognitions task. The authorship of four authors were evaluated: Arthur C. Doyle, Bram Stoker, Charles Darwin and Hector H. Munro (Saki).}
\end{table*}

\begin{figure*}[!htbp]
\begin{center}
       \includegraphics[width=0.93\linewidth]{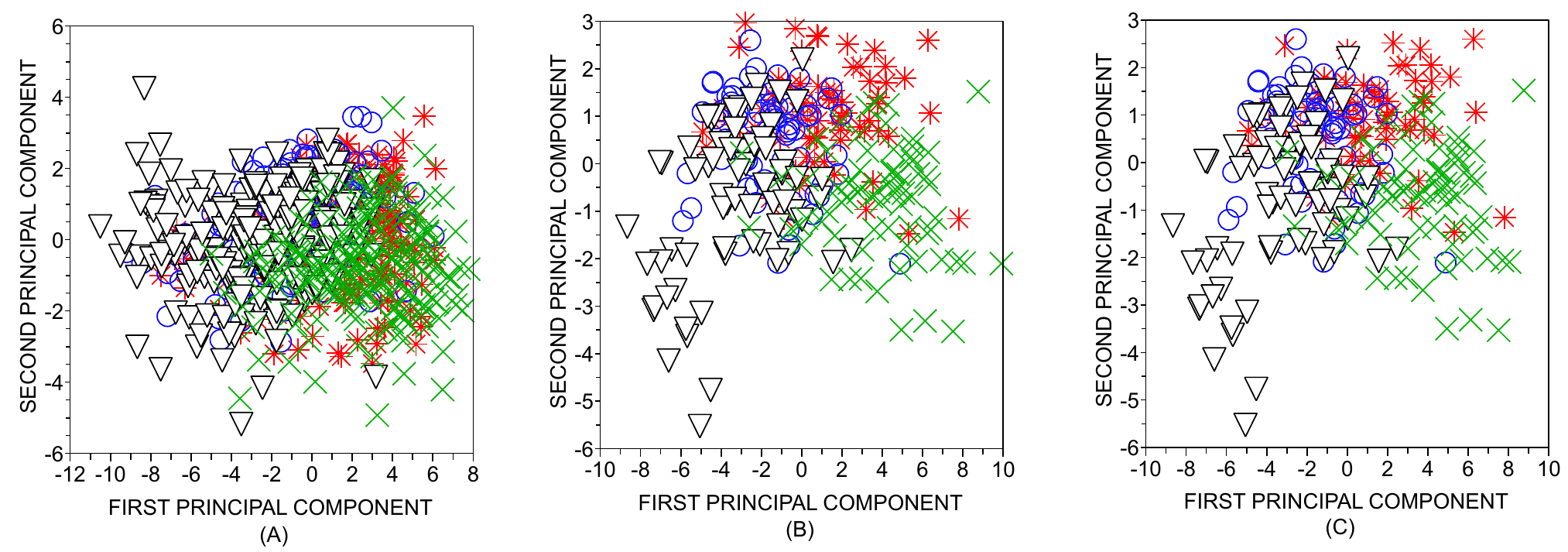}
        \caption{\label{pca_reduced}Visualization of the projection of the eleven attributes employed for the classification of authorship. The principal component analysis was employed. The length of the subtexts considere for the purpose of authorship recognition were: (a) $W = 500$; (b) $W = 1,000$; and (c) $W = 1,500$. Note that the discriminability increases as the subtexts become larger.
        }
	\end{center}
\end{figure*}

\begin{figure*}[!htbp]
\begin{center}
    \includegraphics[width=0.67\linewidth]{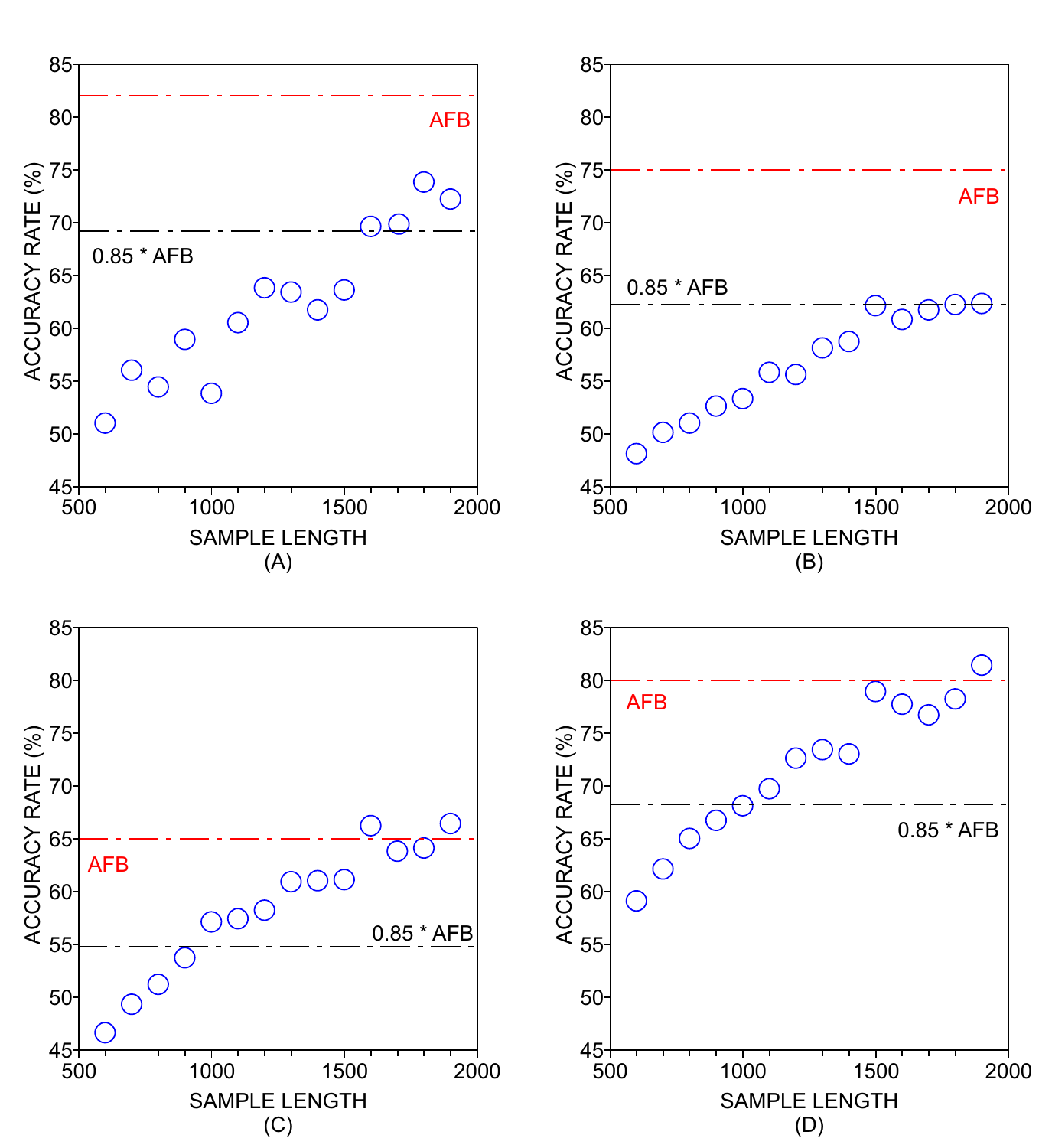}
        \caption{\label{ddet}\Red{Accuracy rate as a function of the subtext length. The classifiers employed were (a) kNN; (b) Naive Bayes; (c) C4.5; and (d) SVM. The accuracy rate obtained with full texts (AFB) is represented as a red dashed line. Considering all the classifiers, one can obtain 85\% of the discriminability found with full books ($W=21,400$) if one analyzes subtexts comprising at least $W = 1,700$ tokens.}}
	\end{center}
\end{figure*}

The performance obtained for the authorship recognition task is shown in Table \ref{tab.2}. A visualization of the classification is provided in Figure \ref{pca_reduced}. The highest accuracy rate found was 86.67\% (p-value $< 10^{-10}$), which confirms that the  topological  characterization  of subtexts is able to discriminate authors. In all classifiers, the lowest accuracy rates were found for the shortest subtexts ($W=500$). This poor discriminability of authors  can be attributed to the very short length employed for the task (see Figure \ref{ddet}). {In order to compare the performance obtained with short and full texts, it is possible to define a threshold $W_L$ from which the accuracy rate surpasses the value $\theta~\times~$AFB, where AFB is the accuracy found with the traditional approach based on full texts. Using $\theta = 0.85$, the following thresholds were obtained: $W_L = 1,600$ (kNN), $W_L = 1,700$ (Bayes), $W_L = 1,000$ (C4.5) and $W_L = 1,000$ (SVM).
Therefore, in the dataset provided in Table \ref{tab.books.recog}, one can obtain 85\% of the discriminability found with full books ($W=21,400$) if one analyzes subtexts comprising at least $W = 1,700$ tokens. }

\begin{table}[h]
\begin{center}
\caption{\label{tab.2}Accuracy rate for the authorship recognition task.}
\begin{tabular}{ccccc}
\hline
 Sample  & {\bf kNN}  & {\bf Bayes} & {\bf C4.5} & {\bf SVM} \\
length   &  (\%) & (\%) & (\%) & (\%)\\
\hline
500    & 49.60  & 46.70 & 48.10 & 56.90\\
1,000   & 53.33  & 53.30 & 57.10& 68.09\\
1,500  & 63.57  & 62.10 & 61.07& 78.92\\
2,000  & 72.50 & 63.50 & 63.50 & 78.50\\
2,500  & 71.25  & 66.87 & 68.75& 81.25\\
3,000  & 68.57  & 67.14 & 59.28& 83.57\\
5,350  & 75.00 & 77.50 & 80.00 & 86.25\\
7,130  & 81.60  & 73.30 & 81.60& 86.67\\
21,400  & 82.00  & 75.00 & 65.00 & 80.00\\
\hline
\end{tabular}
\end{center}
\centering
\Red{Variation of accuracy rate  as a function of subtext length $W$.  The best accuracy rate was found for the SVM, when  the sample length $W = 7,130$ tokens}.
\end{table}

If one compares the accuracy rates found for distinct values of $W$, it is interesting to note that the highest accuracy rates does not occur when full books are used. As a matter of fact, the highest accuracy rates for kNN, Bayes, C4.5 and SVM were obtained for $W=21,400$, $W = 5,350$, $W = 7,130$ and $W = 5,350$, respectively. This observation suggests that, when a suitable sampling is performed, the performance of the classification can even be improved. In line with the results reported in the literature~\cite{systematic}, the SVM outperformed other classifiers (for a given subtext length). In particular, this classifier yielded high accuracy rates even for short subtexts ($W=2,500$). The accuracy obtained for the SVM when $W=3,000$ even surpassed all the accuracies found for the other classifiers. This means that the combination of local characterizations via sampling with support vector machines outperformed the classification based on full books. Given these observations, it seems that support vector machines are more robust than other classifiers when shorter texts are analyzed.
%
%

\section*{Conclusions}

This study probed the influence of sampling texts in the topological analysis of word adjacency networks. An individual analysis of variability of each network measurement revealed that most of them display a low variability across samples. The only exceptions were the skewness of the average shortest path length and the skewness of the intermittency, as revealed by high values of variability across samples even for larger subtexts. Taken together, these results evidence that short pieces of texts are suitable for network analysis, because the sampling process yields weakened statistical fluctuations for short texts.

The influence of the sampling process on a practical classification task was also investigated. Surprisingly, high accuracy rates could be found for texts comprising $1,700$ tokens, which amounts to less than 8\% of the length of a full book. The SVM classifier turned out to be the best classifier for the authorship recognition task based on short texts, as it outperformed the other three traditional classifiers. These results confirm that,
when sufficiently large texts are generated, the sampling does not significantly affect the performance of the classification. Actually, the local characterization of texts might even improve the performance of the classifiers. \Red{In addition to allowing the use of short texts in classification tasks based on stylometry via topological characterization of word adjacency networks, the use of small pieces of texts tends to reduce the effects of the so-called curse-of-dimensionality~\cite{duda}, as more training examples are included in the attribute space.} \Red{A possible weakness of the sampling method is that it can only be applied to large texts. The sampling of short documents generates texts with high topological variability. In these cases, other models should be used to capture relevant features for the specific task~\cite{Manning}. The analysis of short texts has been performed, for example, to detect sentiments. When short comments in product reviews or tweets are analyzed, semantical methods methods have been applied~\cite{newvenues,science}}.

This paper showed, as a proof of principle, that smaller pieces of texts can also be useful in textual network analysis. Following this research line, future works could, for example, apply the techniques described here to identify stylistic inconsistencies in written texts. Such inconsistencies could be found, for example, by identifying topological outliers, i.e. subtexts whose topology is different from other observations in the same book.
The identification of stylistic inconsistencies could also be useful for recognizing multiple authorship or even plagiarisms~\cite{plagio}. Most importantly, the techniques described here could also be extended in a straightforward fashion to study written texts as temporal series~\cite{dnjp}, thus allowing the study of texts as time-varying complex networks~\cite{time}.


\ack
I acknowledge financial support from S\~ao Paulo Research Foundation (FAPESP-Brazil) (grant number 13/06717-4).

\newpage

\setcounter{section}{1}

\end{document}